\documentclass[10pt,twocolumn,letterpaper]{article}
\ifdefined\pdfminorversion\pdfminorversion=7\fi

\usepackage{wacv}

\providecommand{\Description}[1]{}

\definecolor{wacvblue}{rgb}{0.21,0.49,0.74}
\usepackage[breaklinks,colorlinks,allcolors=wacvblue]{hyperref}
\usepackage{balance}
\usepackage{orcidlink}
\hypersetup{
  pdftitle={SketchBloom: Training-Free Vector Sketch Outpainting with Stroke-Level Style Control},
  pdfauthor={Quang-Thinh Nguyen, Duy-Khang Do, Tam V. Nguyen, Minh-Triet Tran, Trung-Nghia Le}
}

\def\wacvPaperID{1745}
\def\confName{WACV}
\def\confYear{2027}

\title{SketchBloom: Training-Free Vector Sketch Outpainting with Stroke-Level Style Control}

\author{%
Quang-Thinh Nguyen$^{1,2\dagger}$\,\orcidlink{0009-0002-1283-8663} \quad
Duy-Khang Do$^{1,2\dagger}$\,\orcidlink{0009-0003-7430-1586} \quad
Tam V. Nguyen$^{3}$\,\orcidlink{0000-0003-0236-7992} \\
Minh-Triet Tran$^{1,2}$\,\orcidlink{0000-0003-3046-3041} \quad
Trung-Nghia Le$^{1,2\ddagger}$\,\orcidlink{0000-0002-7363-2610} \\[2mm]
{\normalsize $^{1}$University of Science, Ho Chi Minh City, Vietnam} \\
{\normalsize $^{2}$Vietnam National University, Ho Chi Minh City, Vietnam} \\
{\normalsize $^{3}$University of Dayton, Ohio, United States} \\[1mm]
{\small $^{\dagger}$Equal contribution \quad $^{\ddagger}$Corresponding author}\\[1mm]
\textbf{Project page: \url{https://khangcntn123.github.io/SketchBloom}}
}

\begin{document}

\twocolumn[{%
  \renewcommand\twocolumn[1][]{#1}%
  \maketitle
  \begin{center}
    \includegraphics[trim={0 0 0 1.3cm},clip,width=0.9\textwidth]{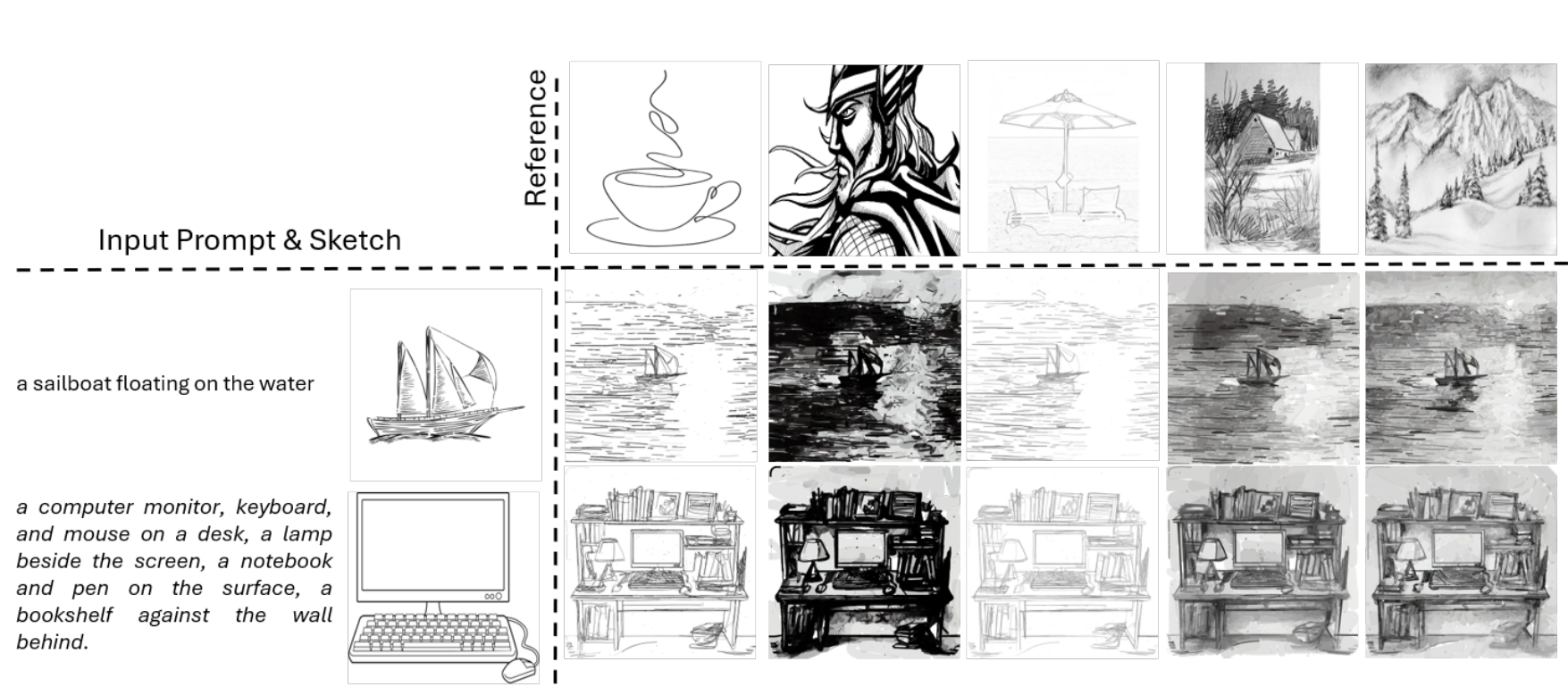}
  \end{center}
  \vspace{-4mm}
  \captionsetup{hypcap=false}
  \captionof{figure}[Overview]{Style-guided vector sketch outpainting with SketchBloom. Given an input sketch, text prompt, and reference style, SketchBloom extends the scene beyond the original boundary while preserving structure and transferring the reference style to generated strokes. Each row shows an input sketch and prompt followed by the resulting SVG outpainting in the corresponding style.}
  \label{fig:teaser}
  \vspace{3mm}
}]


\begin{abstract}
Outpainting sparse sketches while preserving contour coherence and stroke style remains challenging. We present SketchBloom, a training-free framework that combines pretrained raster outpainting and style-transfer guidance with per-instance vector optimization. Given a sketch, a scene prompt, and an optional reference style, it produces an editable Scalable Vector Graphic (SVG) composed of B\'ezier strokes. Adaptive latent blending balances content contours and reference texture without model retraining. A reference-derived stroke vocabulary initializes contour- and DINO-attention-guided strokes, which are optimized separately for structure and texture before contour-emphasizing multiply fusion. As in prior optimization-based vector sketching, no task-specific network is trained; our contribution is the synthesis design specialized to sketch outpainting. On LineArtOut-1K, comprising 1,024 input and style combinations, SketchBloom improves perceptual, semantic, and structural fidelity over the evaluated vector baselines when all methods vectorize the same outpainted raster. This protocol evaluates vector synthesis rather than end-to-end outpainting or reference-style superiority. Runtime is approximately six minutes per pair on a single NVIDIA H200 GPU.
\end{abstract}



\section{Introduction}

Sketching conveys structural intent and stylistic identity through sparse strokes. Although diffusion-based outpainting extends photorealistic images effectively~\cite{rombach2022high,song2020denoising}, extending line art requires generated content to preserve both coherent contours and the drawing's stroke vocabulary.

Photographic style-transfer methods~\cite{DBLP:conf/cvpr/ChungHH24,10657939} can blur contour-dominant line art into grayscale approximations. Stroke-based methods~\cite{vinker2022clipasso,xing2023diffsketcher} primarily address generation from scratch or raster vectorization, with limited support for extending sketches under an independent style reference. Raster generators~\cite{rombach2022high,zhang2023adding,ye2023ip} also lack stroke-level editing and resolution independence. These limitations motivate combining diffusion semantics with explicit vector synthesis.

We present \textbf{SketchBloom}, a training-free framework for style-guided sketch outpainting that produces SVGs composed of optimized B\'ezier curves (\cref{fig:teaser}). Per-instance stroke optimization is an established paradigm~\cite{DBLP:journals/tog/LiLGR20,DBLP:conf/nips/FransSW22,vinker2022clipasso}; our contribution lies in adapting initialization, objectives, and composition to this task.

Given a sparse sketch, a scene prompt, and an optional reference style, ControlNet~\cite{zhang2023adding} first extends the content. Texture-aware style transfer blends DDIM-inverted~\cite{song2020denoising} content and edge features with reference features to balance structure and texture. A hybrid initializer then combines contour-guided warping with DINO-attention-guided placement~\cite{9709990}, using a reference-derived stroke vocabulary. Separate DiffVG~\cite{DBLP:journals/tog/LiLGR20} optimization passes align structure strokes to contours and texture strokes to the stylized raster, followed by multiply fusion. Our ablation shows improved convergence over random initialization.

We introduce \textbf{LineArtOut-1K}: 47 base sketches yield 64 outpainted content images, each paired with 16 styles, giving 1,024 combinations. Experiments and a user study support improved vectorization fidelity with outpainted content held fixed. This protocol assesses downstream vector outputs; it does not establish superiority in end-to-end outpainting or reference-style transfer.

Our contributions are:
\begin{itemize}
    \item A vector sketch outpainting framework coupling replaceable raster guidance with differentiable B\'ezier optimization. Hybrid initialization transfers reference width, color, opacity, and template geometry while combining contour and semantic cues.
    \item Decoupled structure and texture optimization: a patch-based multi-scale contour objective preserves sparse boundaries, while a global reconstruction objective captures tonal appearance.
    \item Deterministic multiply fusion and the LineArtOut-1K benchmark. Controlled comparisons, ablations, a user study, runtime analysis, and a raster-generator-plus-vectorizer comparison characterize quality and practical trade-offs.
\end{itemize}

\section{Related work}

\paragraph{Vector sketch generation.}
DiffVG~\cite{DBLP:journals/tog/LiLGR20} enables differentiable rasterization, supporting per-instance stroke optimization with CLIP supervision~\cite{DBLP:conf/nips/FransSW22,vinker2022clipasso,10378256}. SketchBloom inherits this training-free paradigm. Later methods incorporate diffusion priors~\cite{xing2023diffsketcher,10658565,jain2023vectorfusion}, stroke pruning~\cite{DBLP:conf/aaai/HirschornJA24}, splatting-based rendering~\cite{liu2025bzier}, and direct raster-to-sketch prediction~\cite{arar2025swiftsketch}. Vector customization extends synthesis to exemplar-preserving editing and style control~\cite{10.1145/3610548.3618232,10.1145/3721238.3730707}. AutoSketch~\cite{10.1145/3757377.3763828} completes partial vector sketches using style descriptions derived from the input. SketchBloom instead conditions on an independent style reference, performs scene-level raster outpainting, and separately optimizes structural and textural strokes before composition.

\begin{figure*}[t!]
    \centering
    \includegraphics[width=\textwidth]{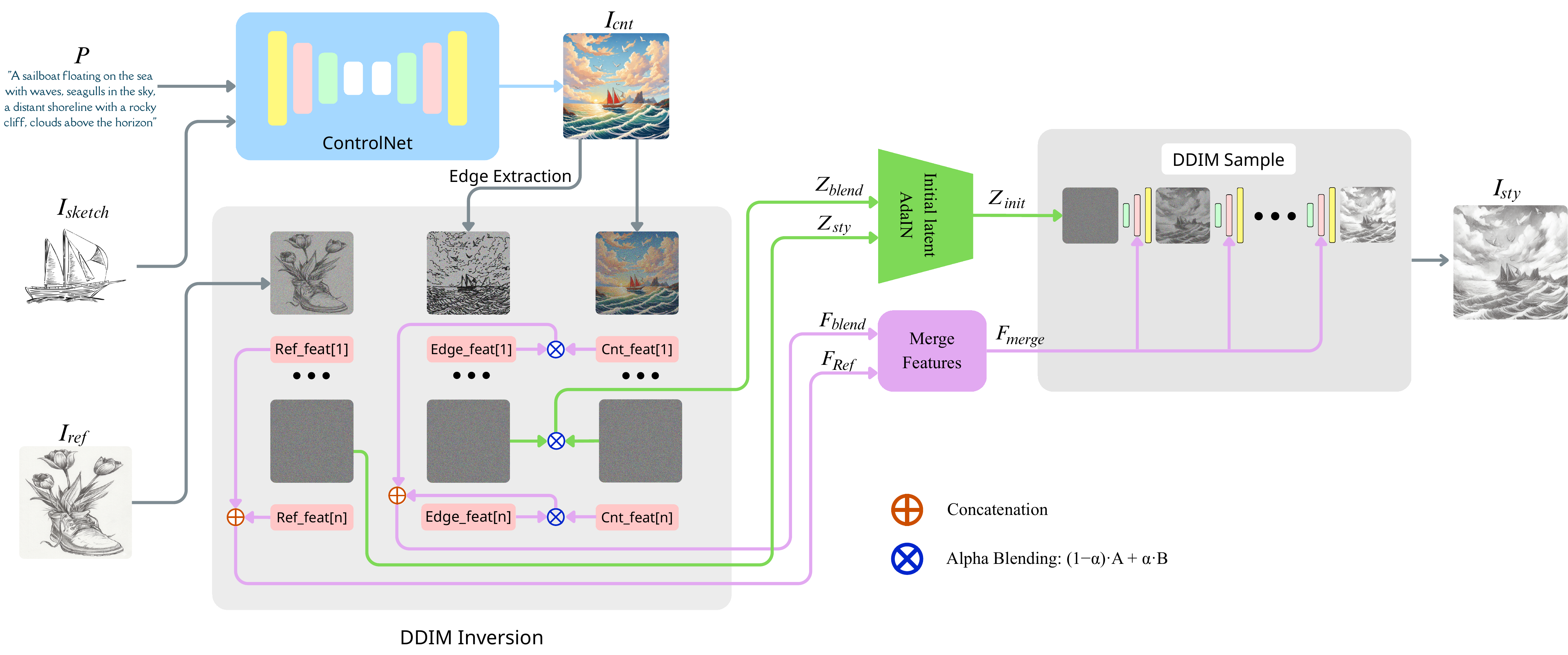}
    \caption{Raster guidance pipeline. ControlNet expands
    $I_{\text{sketch}}$ into $I_{\text{cnt}}$, from which
    $I_{\text{edge}}$ is extracted. DDIM inversion, adaptive latent blending,
    and style-sourced attention features combine content structure with
    $I_{\text{ref}}$, producing the stylized target $I_{\text{sty}}$.}
    \Description{A horizontal pipeline diagram 
on a white background. On the far left, 
three inputs are arranged vertically: a 
text prompt box with italicised sailboat 
scene description, a small sailboat line 
sketch labelled I\_sketch below the prompt, 
and a pencil sketch of tulips in a leather 
boot labelled I\_ref at the bottom. Both 
sketch inputs feed leftward into a large 
blue ControlNet block containing coloured 
vertical bar icons, which outputs a vivid 
painted sailboat image labelled I\_cnt at 
the top right of the block. A grey rounded 
rectangle labelled DDIM Inversion occupies 
the centre and contains three vertical 
columns of stacked grey feature-map 
thumbnails connected by orange plus and 
blue cross-circle operator icons. Green 
horizontal arrows carry Z\_blend and 
F\_blend rightward; purple arrows carry 
Z\_sty and F\_sty. On the right, a large 
grey rounded rectangle labelled DDIM 
Sample contains a row of alternating grey 
and coloured vertical bar icons. Two 
branches feed into it from the left: a 
green funnel-shaped AdaIN block producing 
Z\_init, and a purple rectangular Merge 
Features block producing F\_merge. The 
final output I\_sty on the far right is a 
grayscale sketch of the sailboat scene 
rendered in the ink-wash style of I\_ref.}
    \label{fig:style_transfer}
\end{figure*}

\paragraph{Sketch style transfer.}
Sparse stroke geometry makes sketch stylization distinct from photographic texture transfer. Gram-based~\cite{7780634} and diffusion-based approaches~\cite{DBLP:conf/cvpr/ChungHH24,10657939,ye2023ip} can blur line structure. Related work explores stroke attributes~\cite{yang2025stroke2sketch}, attention-guided optimization~\cite{Zhou_2025_CVPR}, reference correspondence~\cite{10.1145/3550454.3555504}, personalized extraction from diffusion features~\cite{yun2024representative}, and joint abstraction and style control~\cite{Lu2026AbstractionIS}. Closest to our raster stage, M3S~\cite{li2025text} generates reference-guided raster sketches from text, injecting auxiliary reference keys and values with linear smoothing and controlling multiple styles through joint AdaIN. Our stage instead restyles an existing outpainted image, blending content and edge latents to protect contours. Its raster output guides vector synthesis, where reference-derived stroke attributes and DINO-guided placement provide explicit stroke-level control.

\section{Proposed method}
\label{sec:method}

\subsection{Overview}
\label{sec:overview}

Given a sparse sketch input $I_{\text{sketch}}$, a text prompt $P$, and a
reference style image $I_{\text{ref}}$, SketchBloom synthesizes a
stylistically faithful SVG extension of the original line art. When no style
reference is provided, $I_{\text{ref}}$ defaults to $I_{\text{sketch}}$
itself, allowing the pipeline to derive stylistic cues directly from the
input sketch. The pipeline proceeds through three sequential stages.

\paragraph{Stage 1: Raster preprocessing.}
The input sketch is outpainted via a ControlNet-conditioned~\cite{zhang2023adding}
diffusion model to produce a semantically extended content image
$I_{\text{cnt}}$, from which structural contours are extracted as edge map
$I_{\text{edge}}$. A StyleID-inspired~\cite{DBLP:conf/cvpr/ChungHH24} style transfer
mechanism then produces the stylized raster $I_{\text{sty}}$ that fuses the
tonal, textural, and colorimetric properties of $I_{\text{ref}}$ onto
$I_{\text{cnt}}$, serving as the perceptual optimization target for the
texture refinement stage (\cref{sec:style_transfer,fig:style_transfer}).

\begin{figure*}[t!]
    \centering
    \includegraphics[width=\textwidth]{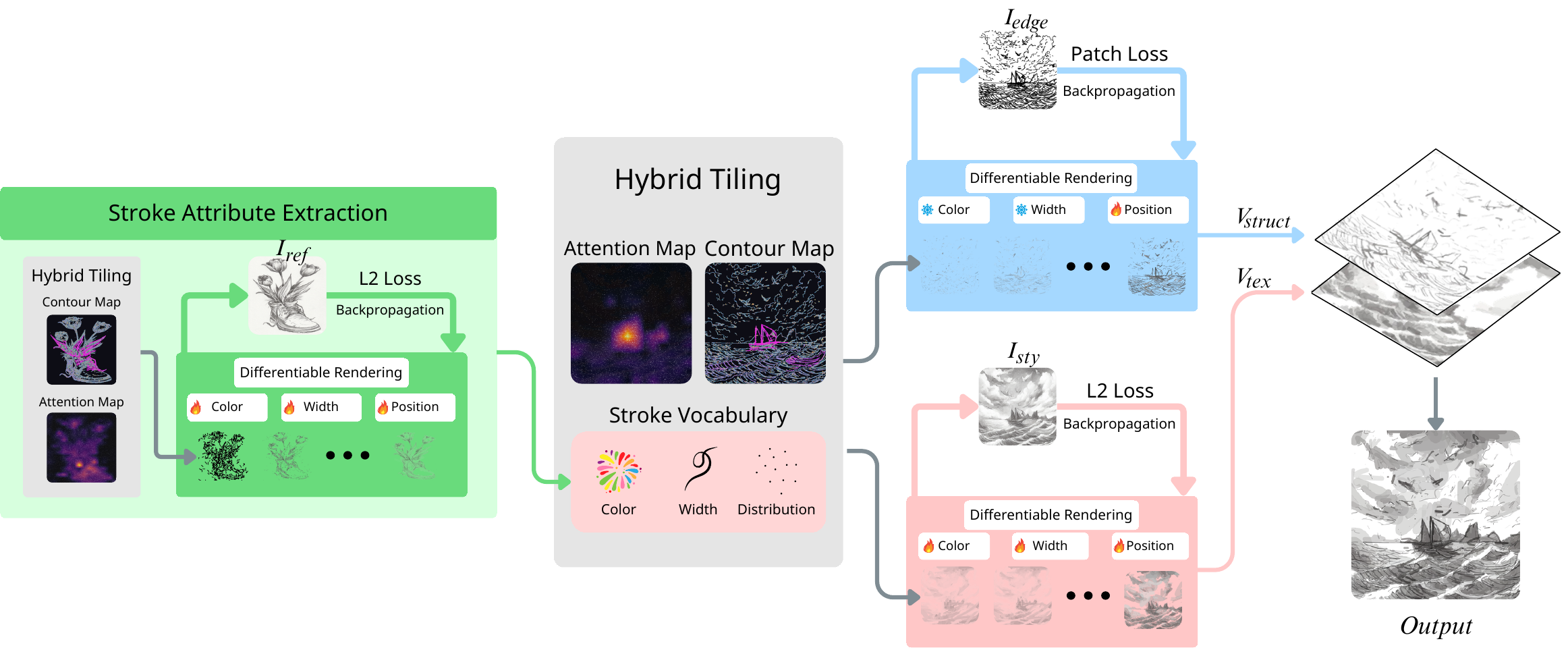}
    \caption{Vector-native synthesis. A reference-derived stroke vocabulary
    initializes two passes: structure optimizes geometry against
    $I_{\text{edge}}$, whereas texture optimizes stroke attributes against
    $I_{\text{sty}}$. Multiply blending places
    $\mathcal{V}_{\text{struct}}$ over $\mathcal{V}_{\text{tex}}$ to produce
    the final SVG $\mathcal{V}^{*}$.}
    \Description{A three-section horizontal 
pipeline diagram. The leftmost green-
bordered block labelled Stroke Attribute 
Extraction contains a purple heat-map 
attention map thumbnail and a dark contour 
map thumbnail feeding into a Hybrid Tiling 
label, alongside I\_ref shown as the 
floral-boot pencil sketch. An $L_2$ Loss 
Backpropagation label sits above a 
Differentiable Rendering bar showing 
three flame icons for Color, Width, and 
Position. A sequence of four increasingly 
refined grey stroke-set thumbnails extends 
rightward. The centre grey-bordered block 
labelled Hybrid Tiling shows the same 
attention and contour map thumbnails above 
a Stroke Vocabulary row with three icons: 
a multicolour paint-wheel for Color, a 
calligraphic stroke glyph for Width, and 
a sparse dot pattern for Distribution. 
Two branches emerge from the centre block. 
The upper blue-bordered block shows I\_edge 
as a dark engraving-style sketch of a 
sailboat with birds, Patch Loss 
Backpropagation, a Differentiable 
Rendering bar with two snowflake icons 
for frozen Color and Width and one flame 
icon for Position, and four stroke-set 
thumbnails ending in V\_struct. The lower 
pink-bordered block shows I\_tex as a 
grayscale tonal image, $L_2$ Loss 
Backpropagation, a Differentiable 
Rendering bar with three flame icons, 
and four stroke-set thumbnails ending 
in V\_tex. On the far right, V\_struct 
and V\_tex converge into a final Output 
label beside a detailed grayscale SVG 
sailboat sketch with visible wave and 
bird strokes.}
    \label{fig:loop}
\end{figure*}

\paragraph{Stage 2: Differentiable stroke initialization and optimization.}
A hybrid tiling strategy combines contour-guided and attention-guided
placement to initialize $N$ B\'ezier strokes per synthesis branch $\{s_i\}_{i=1}^{N}$ over the
canvas (\cref{sec:init,fig:loop}). These strokes are refined through a
structured optimization loop that separately aligns stroke geometry to
$I_{\text{edge}}$ and stroke appearance to $I_{\text{sty}}$, producing a
structure stroke set $\mathcal{V}_{\text{struct}}$ and a texture stroke set
$\mathcal{V}_{\text{tex}}$ (\cref{sec:optim,fig:loop}).

\paragraph{Stage 3: Vector fusion.}
$\mathcal{V}_{\text{struct}}$ and $\mathcal{V}_{\text{tex}}$ are composited
into a unified scalable SVG via multiply-blend layering, where dark
structural contours selectively suppress the underlying texture while light
regions remain visually unchanged
(\cref{sec:fusion,fig:loop}).

\subsection{Raster preprocessing}
\label{sec:style_transfer}

The raster preprocessing pipeline produces the stylized raster $I_{\text{sty}}$
through two sequential steps: sketch outpainting and texture-aware style transfer (\cref{fig:style_transfer}).

\paragraph{Sketch outpainting.}
The implementation supports independently configurable horizontal and vertical canvas-expansion factors. For all benchmark experiments, we adopt a fixed symmetric setting with $r_x=r_y=3$ and centrally place $I_{\text{sketch}}$ on the expanded white canvas before resizing it to $1024\times1024$ pixels. This standardized protocol ensures that every compared method receives the same available outpainting area. The expanded sketch is used as the control signal for ControlNet~\cite{zhang2023adding} paired with an SDXL-based Stable Diffusion~\cite{rombach2022high} checkpoint (model identifiers are listed in the supplementary material), conditioned on prompt $P$ for 50 denoising steps with guidance scale $7.5$ and conditioning scale $1.0$. The resulting content image $I_{\text{cnt}}$ is processed by an edge detector to obtain $I_{\text{edge}}$ for downstream vector stages.

\paragraph{Texture-aware style transfer.}
SketchBloom adopts a feature-injection mechanism inspired by StyleID~\cite{DBLP:conf/cvpr/ChungHH24}
to transfer the tonal and textural properties of $I_{\text{ref}}$
into $I_{\text{cnt}}$. When $I_{\text{ref}}$ is not provided, it defaults to
$I_{\text{sketch}}$. All three images (\textit{i.e.}, $I_{\text{cnt}}$, $I_{\text{edge}}$,
and $I_{\text{ref}}$) are independently inverted via DDIM
inversion~\cite{song2020denoising}, yielding terminal latents
$z_{\text{cnt}}$, $z_{\text{edge}}$, $z_{\text{sty}}$ and per-timestep
self-attention feature sequences
$\mathcal{F}_{\text{cnt}}^{(i)}$, $\mathcal{F}_{\text{edge}}^{(i)}$,
$\mathcal{F}_{\text{sty}}^{(i)}$.

\paragraph{Adaptive latent blending.}
Rather than initializing generation directly from $z_{\text{cnt}}$, the
content and edge latents are linearly interpolated as: $z_{\text{blend}} = (1 - \alpha)\, z_{\text{edge}} + \alpha\, z_{\text{cnt}}$, where $\alpha \in [0,1]$ controls the structure-texture trade-off:
$\alpha \to 0$ biases generation toward sharp contours derived from
$I_{\text{edge}}$, while $\alpha \to 1$ retains richer tonal variation from
$I_{\text{cnt}}$. We empirically set $\alpha=0.5$ to balance structural
fidelity and stylistic richness. The same interpolation is
applied to the corresponding feature sequences across all timesteps: $\mathcal{F}_{\text{blend}}^{(i)} =(1 - \alpha)\mathcal{F}_{\text{edge}}^{(i)}+ \alpha\mathcal{F}_{\text{cnt}}^{(i)}$.

Prior to denoising, AdaIN~\cite{8237429} aligns the channel-wise statistics of
$z_{\text{blend}}$ to those of $z_{\text{sty}}$: $z_{\text{init}} = \mathrm{AdaIN}(z_{\text{blend}},\; z_{\text{sty}}),$ ensuring the generation starting point already reflects the reference style
distribution.

\paragraph{Feature assembly and injection.}
At each denoising timestep $i$, the blended feature sequence
$\mathcal{F}_{\text{blend}}^{(i)}$ and the style feature sequence
$\mathcal{F}_{\text{sty}}^{(i)}$ are assembled into a merged feature map
$\mathcal{F}_{\text{merge}}^{(i)}$ according to the following routing strategy.
The key $K$ and value $V$ tensors of each UNet self-attention layer are always
sourced from $\mathcal{F}_{\text{sty}}^{(i)}$, transferring the textural
statistics of $I_{\text{ref}}$. The query $Q$ transitions from style-sourced to blend-sourced after a
predefined threshold timestep $T_q$, allowing early denoising steps to be
globally guided by the reference style while later steps recover
content-consistent spatial structure.

The self-attention output is further blended as:
\begin{equation}
    O^{(i)} = (1 - w_i)\, O^{(i)}_{\text{blend}} + w_i\, O^{(i)}_{\text{sty}},
\end{equation}
where $w_i$ follows a piecewise schedule: it remains constant at $w_0$ for
$i \leq T_w$, then decays via the quintic smoothstep
\begin{equation}
    f(t) = 6t^5 - 15t^4 + 10t^3, \qquad t \in [0,1],
    \label{eq:smoothstep}
\end{equation}
as $w_i = w_0\,\bigl(1 - f\!\left(\tfrac{i - T_w}{I - T_w}\right)\bigr)$
for $i > T_w$, where $I$ is the total number of denoising steps.
This allows style injection to remain strong during early denoising before
progressively relaxing in the final steps to recover content structure.
The assembled $\mathcal{F}_{\text{merge}}^{(i)}$ is injected into the UNet
at each active timestep $i$ during DDIM sampling from $z_{\text{init}}$,
yielding the stylized raster $I_{\text{sty}}$ that serves as the optimization
target for the texture synthesis stage.

\subsection{Stroke initialization via hybrid tiling}
\label{sec:init}

Given the content image $I_{\text{cnt}}$ and its edge map $I_{\text{edge}}$,
SketchBloom constructs an initial stroke set $\{s_i\}_{i=1}^{N}$ through a
hybrid tiling strategy that combines contour-guided placement and
attention-guided sampling. The per-branch stroke budget $N$ is partitioned into three groups: a majority fraction 
$\rho \in [0,1]$ is allocated to contour-guided strokes,
of which two-thirds are placed directly along detected contours and the
remaining third are generated by interpolating between adjacent pairs;
the rest are placed via attention-guided sampling.

\paragraph{Stroke vocabulary.}
Prior to hybrid tiling, a proxy stroke set is initialized directly from
$I_{\text{ref}}$ via SLIC superpixel segmentation~\cite{6205760}.
Each segment yields one stroke: the two farthest convex-hull border points
define the stroke endpoints $(s_i, e_i)$, the midpoint $c_i = (s_i+e_i)/2$
serves as the Bézier control point, and the mean segment color gives stroke
color $\mathbf{c}_i$. Stroke width $w_i$ is estimated as the average
perpendicular distance from each border point to the axis
$\overrightarrow{s_i e_i}$:
\begin{equation}
    w_i = \frac{1}{|\mathcal{B}_i|}
          \sum_{p \in \mathcal{B}_i}
          \frac{\bigl|\overrightarrow{s_i e_i} \times \overrightarrow{s_i p}\bigr|}
               {\|\overrightarrow{s_i e_i}\|},
\end{equation}
where $\mathcal{B}_i$ denotes the set of convex-hull border points of segment
$i$. This proxy set is then optimized against $I_{\text{ref}}$ via
differentiable rendering to refine all attributes, yielding a stroke
vocabulary that captures the visual character of the reference style.
Without this step, placed strokes would carry no stylistic information, as
their color, width, and geometry would be initialized randomly with
$\mathbf{c}_i \sim \mathcal{U}(0,1)^3$ and positions drawn from a random
walk of step radius $0.05$ scaled to canvas size.

\paragraph{Contour-guided placement.}
Edge contours are extracted from $I_{\text{edge}}$ by thresholding and
connected-component analysis, then weighted proportionally to their arc
length for stroke distribution. For each placement position along a contour,
a template stroke is retrieved from the vocabulary via spatial grid lookup:
the canvas is partitioned into a $G \times G$ grid, and the nearest reference
strokes to the query position are sampled as candidates.

The selected template's control points are scaled and warped to conform to
the local contour geometry. Each control point is projected onto the
template's principal axis~$\mathbf{u}_k$ (the leading right singular vector
of the mean-centered control points) to obtain a normalized axial coordinate $t_j\in[0,1]$. Let $\bar{p}_j = p_j - \bar{p}$ be the
centered coordinate of control point~$j$, and let
$\mathbf{r}_j = \bar{p}_j - (\bar{p}_j \cdot \mathbf{u}_k)\,\mathbf{u}_k$
be its perpendicular residual. Let $\mathcal{C}_k(t)$ denote the selected target contour segment and $\boldsymbol{\tau}(t)=(\tau_x(t),\tau_y(t))$ its unit tangent. The unit normal is $\mathbf{n}(t_j)=(-\tau_y(t_j),\tau_x(t_j))$. The warped control point is:
\begin{equation}
    \tilde{p}_j
    \;=\;
    \mathcal{C}_k(t_j)
    \;+\;
    \beta\;
    \mathrm{sgn}(\mathbf{r}_j \cdot \mathbf{n}(t_j))\;
    \|\mathbf{r}_j\|\;
    \mathbf{n}(t_j),
    \qquad \beta = 0.3,
    \label{eq:warp}
\end{equation}
which re-projects the perpendicular residual from the template's local frame
onto the contour normal direction, preserving stroke shape while adapting to
the local geometry of~$\mathcal{C}_k$. Additional strokes are interpolated
between adjacent placed pairs using a quintic smoothstep schedule
(\cref{eq:smoothstep}) to ensure smooth attribute transitions; pairs
whose dissimilarity in color, width, and position falls below $\tau = 0.3$
are skipped to avoid redundancy.

\paragraph{Attention-guided placement.}
For semantically salient regions not covered by detected contours, strokes
are sampled from an attention probability map $\mathbf{A}$ derived from the
$[\text{CLS}]$ token self-attention of a pretrained DINO
ViT~\cite{9709990}. Attention weights from the final transformer block are
averaged across heads and upsampled to canvas resolution. The edge confidence
map $\mathbf{E}$ is obtained analogously from raw contour values.
Both maps are independently softened via a power transform with exponents
$\gamma_{\mathbf{A}} = 0.5$ and $\gamma_{\mathbf{E}} = 0.7$ respectively,
to soften concentrated responses and broaden stroke coverage while preserving the ordering of positive responses.

When the spatial prior is enabled, each map is linearly blended with a
spatial density map $\mathbf{D}$ derived from the proxy stroke set extracted
on $I_{\text{ref}}$ (\cref{sec:init}, ``Stroke vocabulary''):
\begin{equation}
    \begin{aligned}
        \mathbf{A}' &= \lambda\,\mathbf{D} + (1-\lambda)\,\mathbf{A},\\
        \mathbf{E}' &= \lambda\,\mathbf{D} + (1-\lambda)\,\mathbf{E},
    \end{aligned}
    \qquad \lambda = 0.5.
\end{equation}
$\mathbf{D}$ encodes the \emph{template} stroke density of the reference
style and acts as a \emph{consolidation prior} rather than a uniformity
prior: regions that are densely covered in $I_{\text{ref}}$ receive
proportionally more sampled strokes, which preserves the global stroke-mass
distribution of the reference. The mixing weight $\lambda$ thus balances
adherence to the reference layout ($\mathbf{D}$) against semantic salience
($\mathbf{A}$, $\mathbf{E}$). Each blended map is renormalized to a
probability distribution before sampling. Each attention-guided stroke is positioned by sampling
a canvas location from the blended map, centering the template at that
location, and applying a random spatial offset for variety. The final pool
$\{s_i\}_{i=1}^{N}$ is the union of contour-placed, interpolated, and
attention-sampled strokes.

\subsection{Stroke optimization and refinement}
\label{sec:optim}

The initialized stroke set $\{s_i\}_{i=1}^{N}$ is refined through two
independent optimization passes (\cref{fig:loop}). Both steps inherit
their stroke-level priors, including width $w$, color $\mathbf{c}$, and
distribution $\mathbf{D}$, from the \emph{stroke vocabulary} computed
in \cref{sec:init} on the proxy stroke set extracted from
$I_{\text{ref}}$. We emphasize that vocabulary extraction is performed
\emph{before} hybrid tiling and is not part of the canvas refinement loop,
since optimizing canvas-placed strokes directly against $I_{\text{ref}}$
would destroy their spatial layout. The two refinement passes below
operate on the canvas stroke set $\{s_i\}_{i=1}^{N}$ produced by hybrid
tiling.

\paragraph{Step 1: Structure optimization.}
With $w$ and $\mathbf{c}$ held fixed, only geometric parameters $\{\mathbf{P}_i\}$
are optimized toward $I_{\text{edge}}$. Since $I_{\text{edge}}$ exhibits highly
uneven spatial density, a global loss produces inconsistent gradients across
stroke locations. We therefore use a patch-based multi-scale objective over a
$G \times G$ grid:
\begin{equation}
\begin{aligned}
    \mathcal{L}_{\text{struct}}
    &= \frac{1}{G^2} \sum_{i=1}^{G}\sum_{j=1}^{G}
    \Bigl[
        \mathcal{L}_{fine}(\hat{I}_{ij}, I^e_{ij}) \\
    &\qquad
        + \mathcal{L}_{coarse}\!\left(
            \downarrow_r \hat{I}_{ij},
            \downarrow_r I^e_{ij}
        \right)
    \Bigr],
\end{aligned}
    \label{eq:struct_loss}
\end{equation}
where $\hat{I}_{ij}$ and $I^e_{ij}$ are the rendered and target crops at
patch $(i,j)$, and $\downarrow_r$ denotes average pooling with factor $r = 4$. We instantiate $\mathcal{L}_{fine}$ as the mean absolute ($\ell_1$) error and $\mathcal{L}_{coarse}$ as the mean squared ($\ell_2$) error, each averaged over the pixels of the corresponding (pooled) patch.
For the fine-scale mean absolute error, the derivative with respect to a scalar rendered value is $\partial\mathcal{L}_{fine}/\partial\hat{x}=\mathrm{sign}(\hat{x}-x)/|\Omega|$, where $|\Omega|$ counts scalar values in the patch. Its magnitude is constant away from zero residual. The coarse term compares average-pooled patches, anchoring the low-frequency distribution of stroke mass alongside fine-scale contour alignment.
The result is the structure stroke set $\mathcal{V}_{\text{struct}}$, whose
geometry conforms to $I_{\text{edge}}$ while retaining the stroke vocabulary
of $I_{\text{ref}}$.

\paragraph{Step 2: Texture optimization.}
A separate stroke set, re-initialized via the hybrid tiling strategy of
\cref{sec:init}, is optimized with all parameters
$\{\mathbf{P}_i, w_i, \mathbf{c}_i\}$ free against $I_{\text{sty}}$:
\begin{equation}
    \mathcal{L}_{\text{tex}} =
    \|\mathcal{R}(\{s_i\}) - I_{\text{sty}}\|_2^2,
\end{equation}
where $\mathcal{R}$ is the DiffVG differentiable rasterizer~\cite{DBLP:journals/tog/LiLGR20}.

Since $I_{\text{sty}}$ distributes tonal and textural variation across the full
canvas, the global $\ell_2$ objective $\mathcal{L}_{\text{tex}}$ provides stable full-canvas convergence
without over-fitting to patch-level artifacts, yielding the texture stroke set
$\mathcal{V}_{\text{tex}}$.

\subsection{Layer-wise multiply fusion}
\label{sec:fusion}

The unified output is defined as the ordered concatenation
$\mathcal{V}^* = \mathcal{V}_{\text{tex}} \;\|\; \mathcal{V}_{\text{struct}}$,
with $\mathcal{V}_{\text{tex}}$ on the lower layer and
$\mathcal{V}_{\text{struct} }$ on the upper layer; the two stroke sets are embedded as independent SVG groups. The SVG is exported as two layers:
$\mathcal{V}_{\text{tex}}$ is rendered under the \texttt{normal} compositing
mode, while $\mathcal{V}_{\text{struct}}$ uses
\texttt{mix-blend-mode: multiply}. For fully opaque overlaps, the structure-group color $c^s_{xy}$ blends with the underlying texture color $c^t_{xy}$ as: $c^*_{xy} = c^{s}_{xy} \odot c^{t}_{xy}$, where $\odot$ denotes element-wise multiplication in $[0,1]^3$. Stroke opacity is handled by SVG alpha compositing.
Dark structural contours thereby suppress the underlying texture locally,
while regions where $c^{s}_{xy} \approx \mathbf{1}$ remain visually unchanged.

\section{Experiments}

\subsection{Benchmark dataset}

LineArtOut-1K contains 1,024 input and style combinations. Its 47 base sketches cover everyday objects, animals, and people with varied stroke density and complexity. Selected sketch and prompt pairs yield 64 outpainted content images: 43 with complex scene prompts and 21 with moderate prompts. Each is paired with 16 reference images spanning scene and object styles. SketchBloom outputs are generated for all 1,024 combinations. The shared-input comparison evaluates 64 content images; the vector baselines do not accept an independent style reference. Hyperparameters remain fixed across all content and styles, without per-instance tuning.

\subsection{Evaluation metrics}
\label{sec:metrics}

Perceptual evaluation uses the no-reference Aesthetic Score~\cite{schuhmann2022improvedaesthetic}, LPIPS~\cite{8578166}, and DreamSim~\cite{fu2023dreamsim}; Laplacian Variance~\cite{Pertuz2013AnalysisOF} measures sharpness. Gram Distance~\cite{7780634} compares reference-style feature statistics, while DINO cosine similarity~\cite{9709990} measures semantic agreement with the experiment-specific target. Structural metrics are SSIM~\cite{1284395}, mean absolute error (MAE), and Edge F1. Hausdorff~\cite{232073} and Chamfer distances~\cite{Barrow1977ParametricCA} compare skeletonized stroke point sets, measuring extreme and average geometric deviations. Each table specifies its targets and applicable metrics. The loss ablation reports LPIPS, SSIM, Hausdorff distance, and Laplacian variance in the text; implementation details are supplementary.

\subsection{Comparison against state-of-the-art methods}
\label{sec:comparison}

\textit{No existing method simultaneously addresses sketch outpainting and explicit stroke-level style control.} Therefore, our evaluation focuses strictly on \textbf{sketch vectorization quality from a given outpainted raster image}. We compare SketchBloom against five baselines, including CLIPasso~\cite{vinker2022clipasso}, CLIPascene~\cite{10378256}, SwiftSketch~\cite{arar2025swiftsketch}, ControlSketch~\cite{arar2025swiftsketch}, and an adapted DiffSketcher~\cite{xing2023diffsketcher}, using the same outpainted content image $I_{\text{cnt}}$ as input for all. DiffSketcher~\cite{xing2023diffsketcher} is adapted by replacing its original text-conditioned optimization target with $I_{\text{cnt}}$ directly, enabling image-guided vector synthesis in place of its text-to-SVG pipeline.

\paragraph{Scope and fairness.}
Sharing $I_{\text{cnt}}$ holds outpainted content fixed, but SketchBloom also uses style-conditioned raster guidance; conditioning and objectives are not identical across methods. The baselines partly operate outside their intended abstraction regime. We interpret \cref{tab:cmp_scene} as downstream vectorization fidelity under this protocol, not as a ranking on their original tasks or an evaluation of outpainting and style-transfer superiority.

\begin{figure}[t!]
    \centering
    \includegraphics[width=\linewidth]{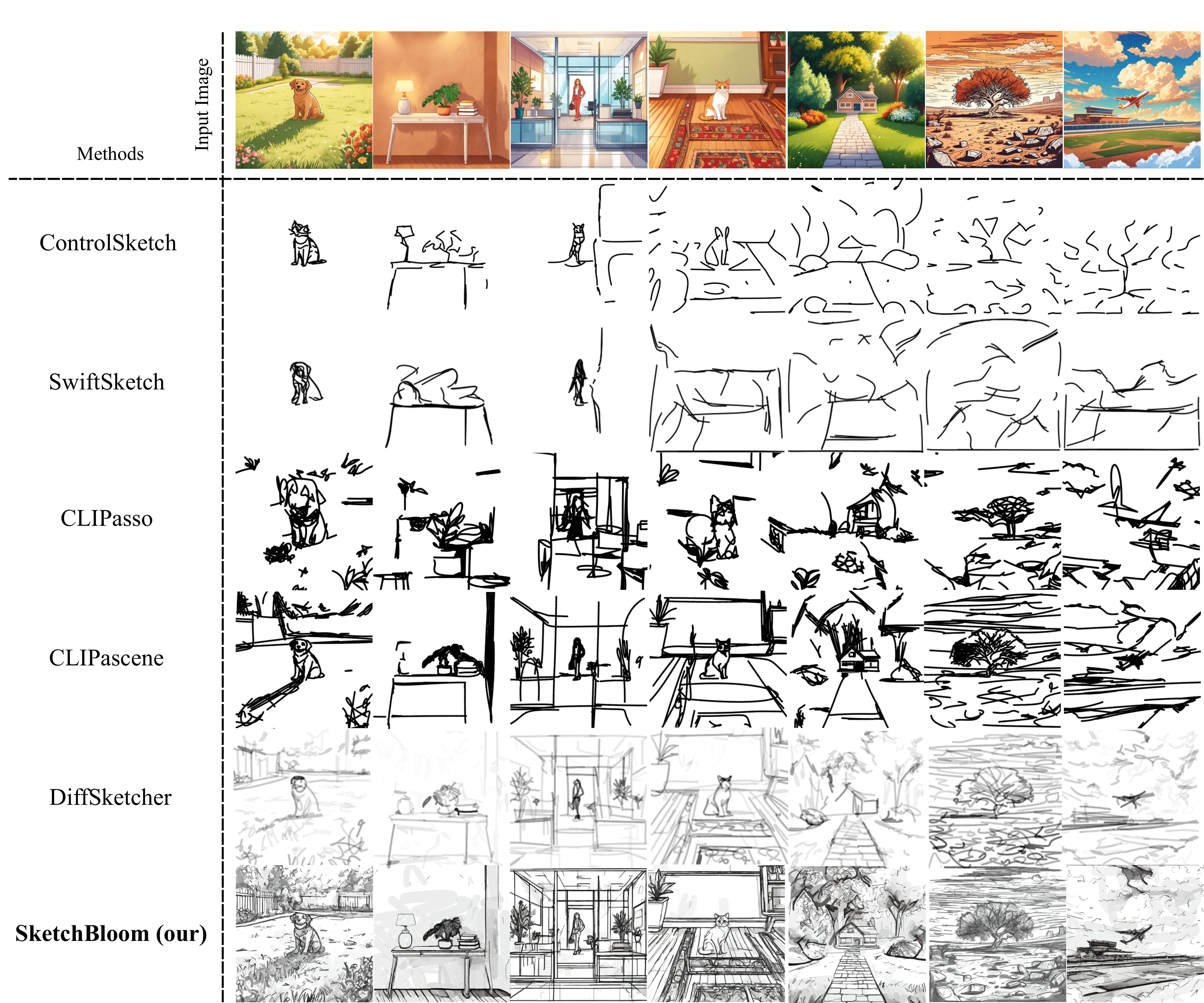}
    \caption{Qualitative comparison under the same raster input.
    SketchBloom (bottom row) better retains object boundaries, perspective,
    and scene detail.}
    \label{fig:cmp_scene}
\end{figure}

Qualitatively, the baselines in \cref{fig:cmp_scene} are either sparse and
disconnected, spatially noisy, over-abstracted, or overly smooth. SketchBloom
instead preserves object contours, perspective cues, and scene depth with a
more coherent stroke layout.

\begin{table}[t!]
\centering
\caption{Quantitative comparison on 64 shared outpainted content images.
All methods receive the same $I_{\text{cnt}}$. DINO and DreamSim use
$I_{\text{cnt}}$ as reference, Edge F1 uses $I_{\text{edge}}$, and
Aesthetic is no-reference. Results are computed per content image. Best values are in \textbf{bold}.}
\label{tab:cmp_scene}
\resizebox{\linewidth}{!}{%
\begin{tabular}{l cccc}
\toprule
\textbf{Method}
  & \textbf{Aesthetic~\cite{schuhmann2022improvedaesthetic}} $\uparrow$
  & \textbf{DINO~\cite{9709990}} $\uparrow$
  & \textbf{DreamSim~\cite{fu2023dreamsim}} $\downarrow$
  & \textbf{Edge F1} $\uparrow$ \\
\midrule
CLIPasso~\cite{vinker2022clipasso}
  & 4.1690 & 0.2064 & 0.7256 & 0.5262 \\
CLIPascene~\cite{10378256}
  & 4.3904 & 0.2022 & 0.7024 & 0.5797 \\
SwiftSketch~\cite{arar2025swiftsketch}
  & 3.9784 & 0.1566 & 0.8305 & 0.3139 \\
ControlSketch~\cite{arar2025swiftsketch}
  & 4.1493 & 0.1470 & 0.7960 & 0.3299 \\
DiffSketcher~\cite{xing2023diffsketcher} (adapted)
  & 4.8020 & 0.3760 & 0.5351 & 0.8325 \\
\textbf{SketchBloom (Ours)}
  & \textbf{5.2186} & \textbf{0.5053} & \textbf{0.4983} & \textbf{0.8866} \\
\bottomrule
\end{tabular}%
}
\vspace{-3mm}
\end{table}

SketchBloom achieves the best value on every reported metric
(\cref{tab:cmp_scene}): Aesthetic $5.2186$, DINO $0.5053$, Edge F1 $0.8866$,
and DreamSim $0.4983$. This indicates stronger perceptual, semantic, and
structural fidelity under the shared-input protocol.

\paragraph{Black-box raster baseline.}
On 24 cases, a black-box generator followed by VTracer obtains Gram Distance/Aesthetic values of $6.401\times10^{-5}/5.662$, versus $1.263\times10^{-4}/5.649$ for SketchBloom's SVG. The native generator achieves $4.900\times10^{-5}/5.813$, so tracing increases Gram Distance and decreases Aesthetic Score. The traced baseline nevertheless better matches dense reference statistics with comparable aesthetics. VTracer represents tonal detail as filled regions; SketchBloom approximates it with a fixed budget of open strokes. This comparison does not isolate representation as the cause of the Gram difference. Full results are supplementary.

\paragraph{Output structure and editability.}
Each SVG contains independent structure and texture groups, each with $N=700$ open B\'ezier strokes storing control points, width, and color. Layers can be hidden or restyled separately, and individual strokes moved or deleted. Reproducing the fused appearance requires support for multiply blending. Quantitative comparisons of path counts, file sizes, and editing effort remain future work.

\paragraph{Runtime and computational cost.}
At $1024\times1024$ with batch size 1 on an NVIDIA H200, runtime is $357.1\pm13.5$\,s per pair. Outpainting and style transfer take $12.6$\,s and $19.4$\,s. Three 500-iteration stages dominate (about $91\%$): vocabulary extraction, structure optimization, and texture optimization take $106.8$\,s, $104.7$\,s, and $113.5$\,s. The reported timings use sequential execution without vocabulary caching. Reference vocabularies could be reused across content images, and the independent synthesis passes could execute concurrently, subject to hardware resources; these speedups have not been measured. This per-instance cost limits practical use; detailed timings are supplementary.

\subsection{Ablation study}

\paragraph{Effect of stroke initialization strategy.}

\begin{figure*}[t!]
    \centering
    \includegraphics[width=0.96\textwidth]{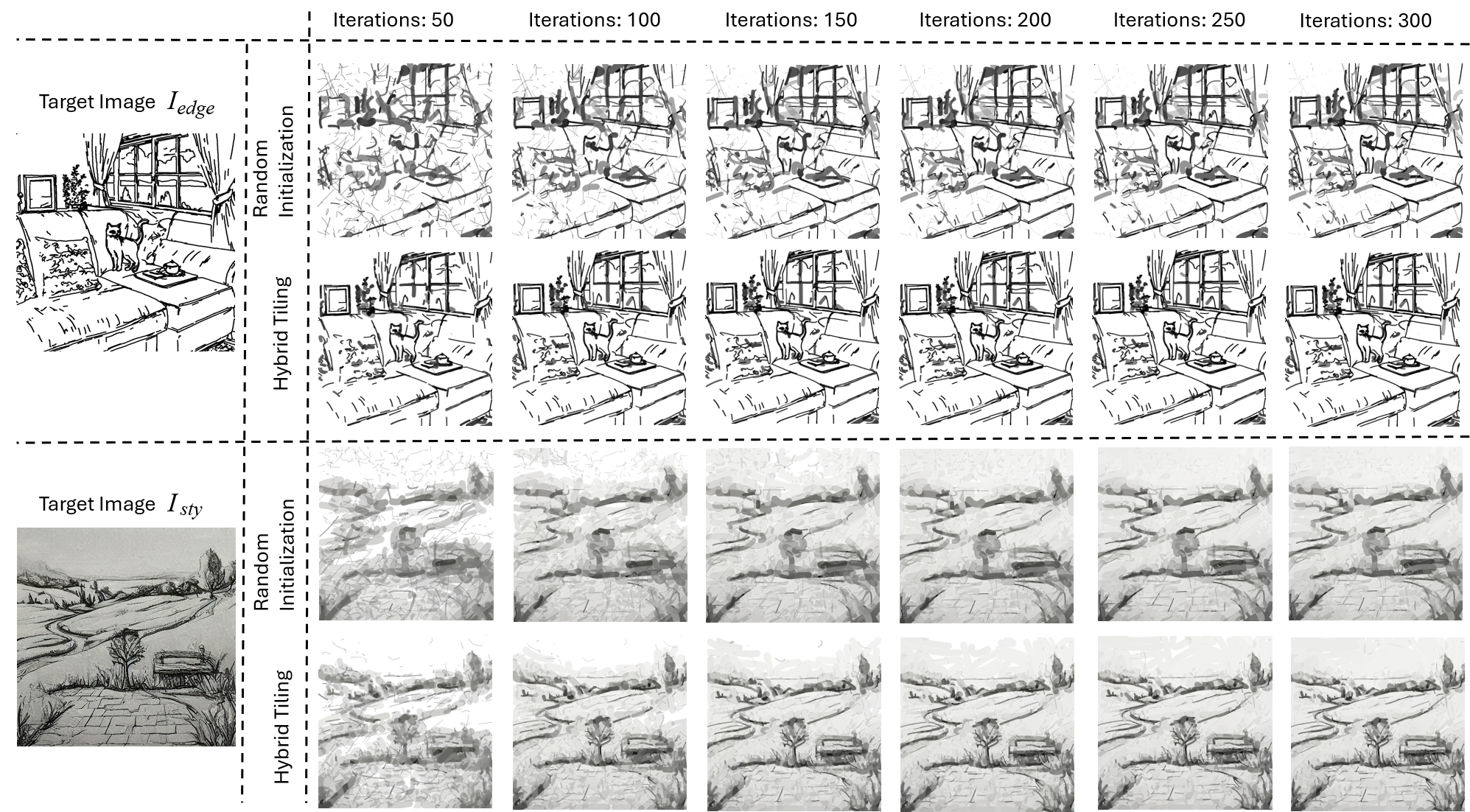}

    \vspace{8pt}
    \begin{minipage}[c]{0.64\textwidth}
        \centering
        \includegraphics[width=\linewidth]{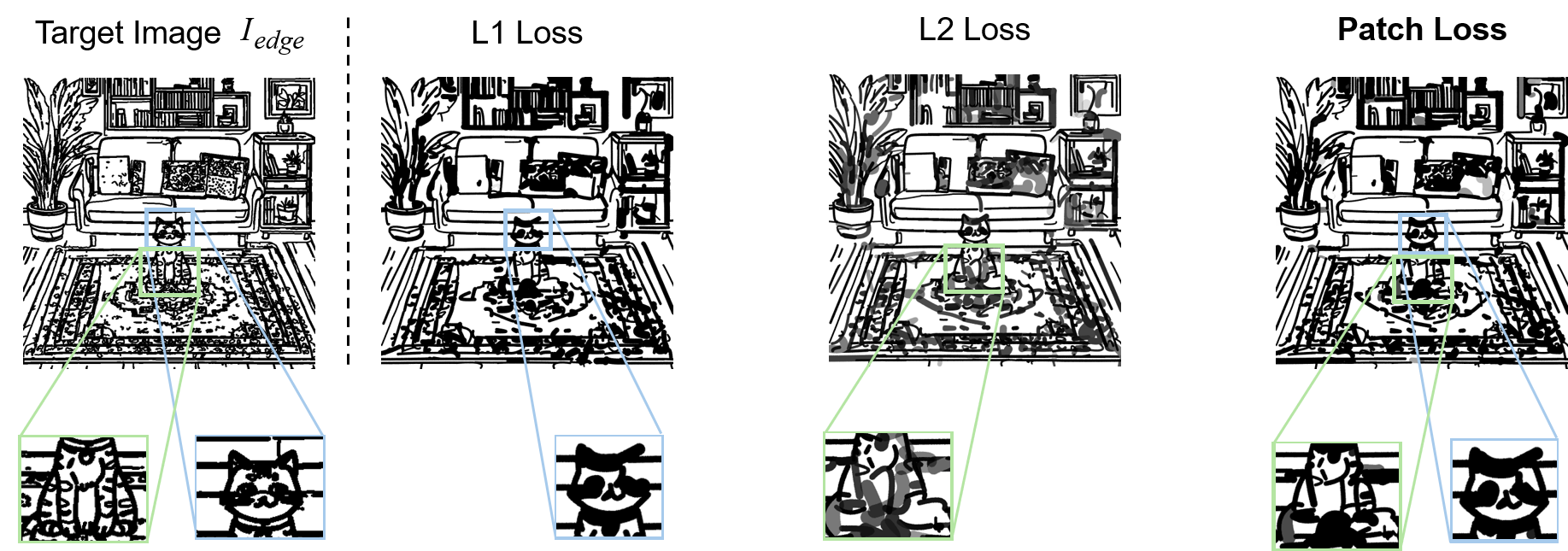}
    \end{minipage}
    \hfill
    \begin{minipage}[c]{0.34\textwidth}
        \centering
        \includegraphics[width=\linewidth]{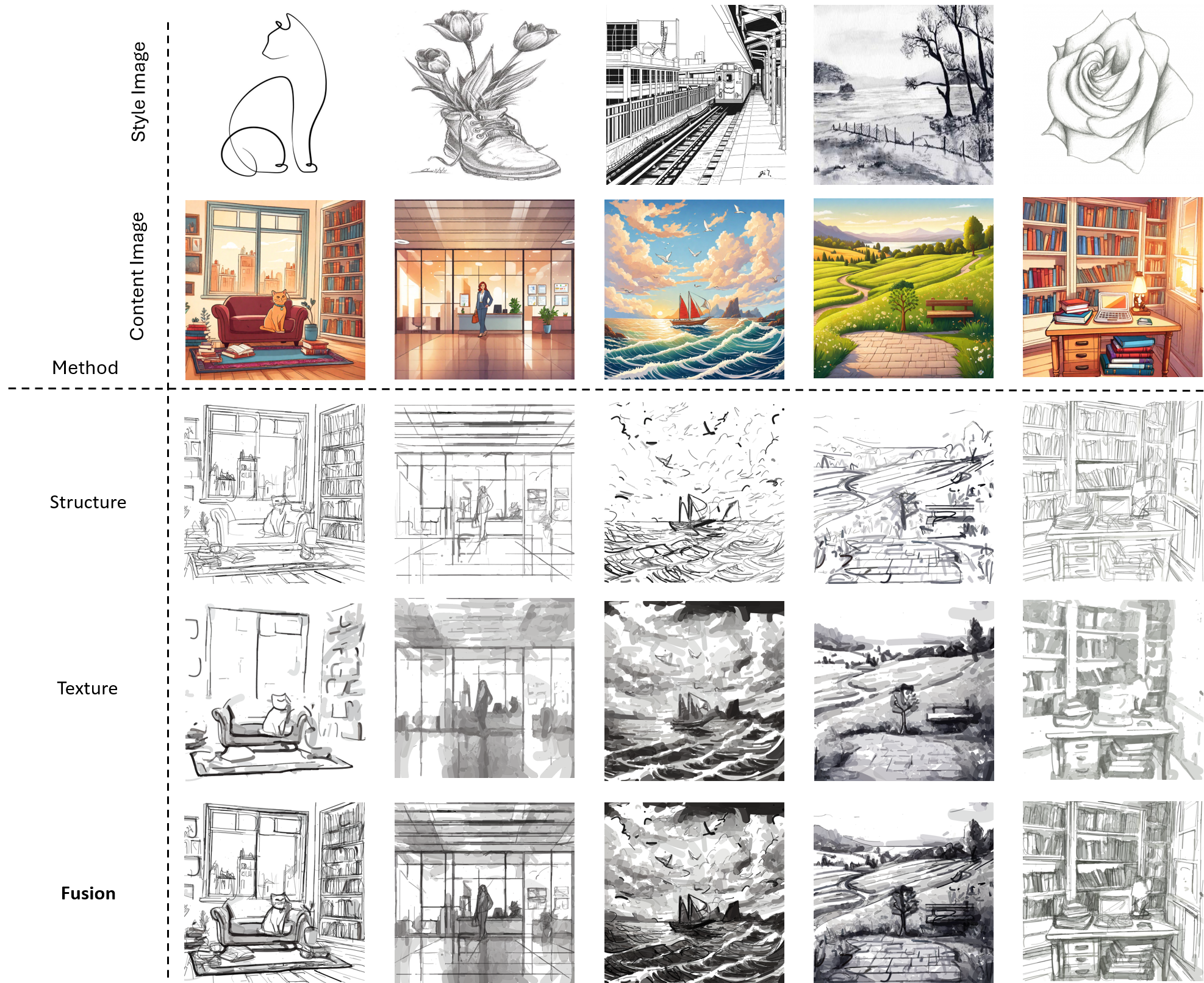}
    \end{minipage}
    \caption{Qualitative ablations. \textit{Top:} hybrid tiling converges
    earlier than random initialization. \textit{Bottom left:} patch loss
    preserves local detail without the gray halos of $L_2$.
    \textit{Bottom right:} multiply fusion combines geometric structure with
    tonal texture.}
    \Description{Three stacked qualitative ablations. The top panel compares
    random and hybrid stroke initialization over six optimization checkpoints
    on structural and stylized targets. The middle panel compares L1, L2, and
    patch losses with magnified local crops. The bottom panel shows five
    style-content pairs followed by structure-only, texture-only, and fused
    vector outputs.}
    \label{fig:ablation_visuals}
\end{figure*}

We compare random placement with the proposed contour- and DINO-guided hybrid initialization. \Cref{tab:abl_init_struct,fig:ablation_visuals} show that hybrid initialization improves all structural metrics and converges earlier under the same optimization budget.

\begin{table}[t!]
\centering
\caption{Structure-only initialization ablation against $I_{\text{edge}}$.
Best values are in \textbf{bold}.}
\label{tab:abl_init_struct}
\resizebox{0.7\linewidth}{!}{
\begin{tabular}{l c c c c}
\toprule
\textbf{Init.}
  & \textbf{MAE} $\downarrow$
  & \textbf{SSIM}~\cite{1284395} $\uparrow$
  & \textbf{Chamfer}~\cite{Barrow1977ParametricCA} $\downarrow$
  & \textbf{Edge F1} $\uparrow$ \\
\midrule
Random
  & 0.4384 & 0.1120 & 0.0006 & 0.7731 \\
\textbf{Hybrid}
  & \textbf{0.2041} & \textbf{0.8068} & \textbf{0.0003} & \textbf{0.8841} \\
\bottomrule
\end{tabular}
}
\end{table}

\paragraph{Effect of layer-wise multiply fusion.}

We validate the vector fusion strategy (\cref{sec:fusion}) against structure-only and texture-only outputs. The proposed variant composites both stroke sets into a two-layer SVG in which structure strokes multiply-blend over texture strokes.

\Cref{tab:abl_fusion,fig:ablation_visuals} show that texture-only obtains the
best Gram Distance but weaker structure, whereas structure-only preserves
geometry but lacks tonal richness. Fusion gives the best Aesthetic, DINO, and
DreamSim scores while visually combining boundary and tonal cues.

\paragraph{Evidence for style fidelity.}
Gram Distance against $I_{\text{ref}}$ (\cref{tab:abl_fusion} and the black-box comparison) shows that texture strokes carry most reference statistics; fusion trades an increase in Gram Distance for contour legibility. We lack a style-specific user study and a broad comparison against reference-guided generators, so qualitative style transfer (\cref{fig:teaser}) should not be read as conclusive style-fidelity superiority.

\begin{table}[t!]
\centering
\caption{Fusion ablation. Gram uses $I_{\text{ref}}$, DINO and DreamSim use
$I_{\text{sty}}$, and Edge F1 uses $I_{\text{edge}}$. Best values are in
\textbf{bold}.}
\label{tab:abl_fusion}
\resizebox{\linewidth}{!}{
\setlength{\tabcolsep}{5pt}
\begin{tabular}{l c c c c c}
\toprule
\textbf{Variant}
  & \textbf{Gram}~\cite{7780634} $\downarrow$
  & \textbf{DINO}~\cite{9709990} $\uparrow$
  & \textbf{DreamSim}~\cite{fu2023dreamsim} $\downarrow$
  & \textbf{Edge F1} $\uparrow$
  & \textbf{Aesthetic}~\cite{schuhmann2022improvedaesthetic} $\uparrow$ \\
\midrule
Texture only
  & $\mathbf{1.82 \times 10^{-4}}$
  & 0.7328
  & 0.1744
  & 0.8067
  & 5.0715 \\
Structure only
  & $2.76 \times 10^{-4}$
  & 0.5800
  & 0.2918
  & \textbf{0.8841}
  & 4.8718 \\
\textbf{Fusion}
  & $2.19 \times 10^{-4}$
  & \textbf{0.7372}
  & \textbf{0.1712}
  & 0.8629
  & \textbf{5.3507} \\
\bottomrule
\end{tabular}
}
\end{table}

\paragraph{Effect of training loss function.}

The patch objective obtains the best Hausdorff distance/Laplacian variance
($22.6493/18{,}852.23$), while $L_2$ gives the best LPIPS/SSIM
($0.1860/0.8284$). The qualitative differences are shown in
\cref{fig:ablation_visuals}.

\subsection{User study}

In a 33-participant 2AFC study, method identities were hidden and output positions randomized. SketchBloom was preferred over CLIPascene by 97.2\%/96.8\%/96.5\%, and over adapted DiffSketcher by 87.4\%/84.9\%/83.5\%, for perceptual quality/content preservation/structural coherence, respectively. Participants were recruited through convenience sampling; the study measures relative perceptual preference under the shared-input protocol and does not evaluate reference-style fidelity. Recruitment, consent, and interface details are supplementary.

\subsection{Failure cases and limitations}

Dense textures can cross semantic boundaries and obscure contours; heterogeneous stroke widths can reduce silhouette coherence. Examples are supplementary. Content-aware stroke allocation and adaptive attribute regularization may mitigate these failures. Hyperparameters (\eg, $\alpha$, $\rho$, $\lambda$, $\beta$, and $G$) are manually selected and fixed across the benchmark. Sensitivity analysis, style-focused user evaluation, and end-to-end outpainting evaluation remain future work.

\section{Conclusion}

We presented SketchBloom, a training-free framework that combines raster
diffusion guidance with differentiable Bézier-stroke optimization for
style-guided vector sketch outpainting. Experiments on LineArtOut-1K,
ablations, and a user study demonstrate improved perceptual, semantic, and
structural fidelity over evaluated vector baselines when vectorizing a shared outpainted raster. Future work will address
dense textures and heterogeneous stroke widths through adaptive stroke
optimization, reduce the per-instance optimization cost, and evaluate style fidelity and editing workflows directly.



\clearpage
{\small
\bibliographystyle{ieeenat_fullname}
\balance
\bibliography{references}
}

\end{document}